\documentclass[10pt,twocolumn,letterpaper]{article}

\usepackage{cvpr}
\usepackage{times}
\usepackage{epsfig}
\usepackage{graphicx}
\usepackage{amsmath}
\usepackage{amssymb}
\usepackage[normalem]{ulem}
\usepackage{afterpage}
\usepackage{booktabs}
\usepackage{color}
\usepackage{verbatim}
\usepackage{soul}

\usepackage[pagebackref=true,breaklinks=true,letterpaper=true,colorlinks,bookmarks=false]{hyperref}

\cvprfinalcopy 

\ifcvprfinal\fi
\begin{document}

\title{Marr Revisited: 2D-3D Alignment via Surface Normal Prediction}

\author{Aayush Bansal \\
Carnegie Mellon University\\
{\tt\small aayushb@cs.cmu.edu}
\and
Bryan Russell \\
Adobe Research\\
{\tt\small brussell@adobe.com}
\and
Abhinav Gupta \\
Carnegie Mellon University\\
{\tt\small abhinavg@cs.cmu.edu}
}

\maketitle
\begin{abstract}

We introduce an approach that leverages surface normal predictions, along with appearance cues, to retrieve 3D models for objects depicted in 2D still images from a large CAD object library. 
Critical to the success of our approach is the ability to recover accurate surface normals for objects in the depicted scene. 
We introduce a skip-network model built on the pre-trained Oxford VGG convolutional neural network (CNN) for surface normal prediction. 
Our model achieves state-of-the-art accuracy on the NYUv2 RGB-D dataset for surface normal prediction, and recovers fine object detail compared to previous methods. 
Furthermore, we develop a two-stream network over the input image and predicted surface normals that jointly learns pose and style for CAD model retrieval. 
When using the predicted surface normals, our two-stream network matches prior work using surface normals computed from RGB-D images on the task of pose prediction, and achieves state of the art when using RGB-D input. 
Finally, our two-stream network allows us to retrieve CAD models that better match the style and pose of a depicted object compared with baseline approaches.

\end{abstract}

\section{Introduction}

Consider the images depicting objects shown in Figure~\ref{fig:teaser_fig}. When we humans see the objects, we can not only recognize the semantic category they belong to, e.g., ``chair'', we can also predict the underlying 3D structure, such as the occluded legs and surfaces of the chair. How do we predict the underlying geometry? How do we even reason about invisible surfaces? These questions have been the core area of research in computer vision community from the beginning of the field. One of the most promising theories in the 1970-80's was provided by David Marr at MIT~\cite{Marr82}. Marr believed in a feed-forward sequential pipeline for object recognition. Specifically, he proposed that recognition involved several intermediate representations and steps. His hypothesis was that from a 2D image, humans infer the surface layout of visible pixels, a 2.5D representation. This 2.5D representation is then processed to generate a 3D volumetric representation of the object and finally, this volumetric representation is used to categorize the object into the semantic category.

While Marr's theory was very popular and gained a lot of attention, it never materialized computationally because of three reasons: (a) estimating the surface normals for visible pixels is a hard problem; (b) approaches to take 2.5D representations and estimate 3D volumetric representations are not generally reliable due to lack of 3D training data which is much harder to get; (c) finally, the success of 2D feature-based object detection approaches without any intermediate 3D representation precluded the need of this sequential pipeline. However, in recent years, there has been a lot of success in estimating 2.5D representation from single image~\cite{Eigen15, Wang15}. 
Furthermore, there are stores of 3D models available for use in CAD repositories such as Trimble3D Warehouse and via capture from 3D sensor devices. 
These recent advancements raise an interesting question: is it possible to develop a computational framework for Marr's theory? In this paper, we propose to bring back the ideas put forth by Marr and develop a computational framework for extracting 2.5D representation followed by 3D volumetric estimation. 

\noindent {\bf Why sequential?} Of course, one could ask why worry about Marr's framework? Most of the available data for training 3D representations is the CAD data (c.f.\ ShapeNet or ModelNet~\cite{Wu15}). While one could render the 3D models, there still remains a big domain gap between the CAD model renders and real 2D images. We believe Marr's 2.5D representation helps to bridge this gap. Specifically, we can train a 2D $\rightarrow$ 2.5D model using RGB-D data, and whose output can be aligned to an extracted 2.5D representation of the CAD models.

Inspired by this reasoning, we used off-the-shelf 2D-to-2.5D models to build our computational framework~\cite{Eigen15, Wang15}. However, these models are optimized for global scene layout and local fine details in objects are surprisingly missing. To overcome this problem, we propose a new skip-network architecture for predicting surface normals in an image. Our skip network architecture is able to retrieve the fine details, such as the legs of a table or chair, which are missing in current ConvNet architectures. In order to build the next stage in Marr's pipeline, we train another ConvNet that learns a similarity metric between rendered CAD models and 2D images using both appearances and surface normal layout. A variant of this architecture is also trained to predict the pose of the object and yields state-of-the-art performance. 

\noindent {\bf Our Contributions:} Our contributions include: (a) A skip-network architecture that achieves state-of-the-art performance on surface normal estimation; (b) A CNN architecture for CAD retrieval combining image and predicted surface normals.  We achieve state-of-the-art accuracy on pose prediction using RGB-D input, and in fact our RGB-only model achieves performance comparable to prior work which used RGB-D images as input.

\begin{figure*}[t]
\centering
\includegraphics[width=\linewidth]{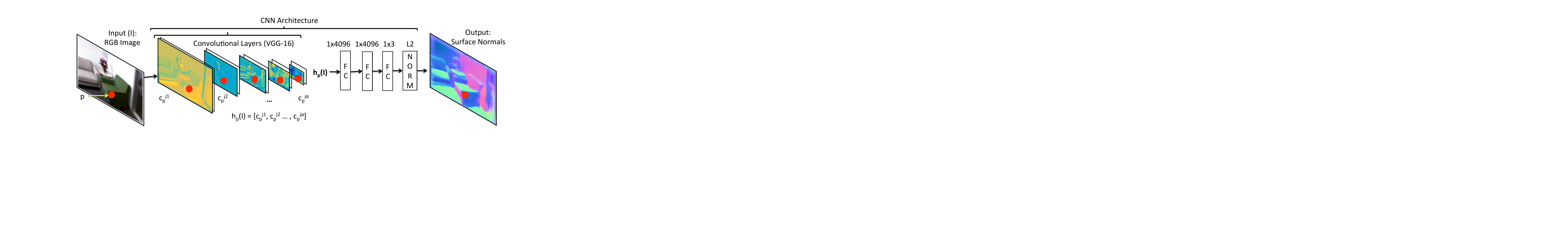}
\caption{
Skip-network architecture for surface normal prediction. 
CNN layer responses are concatenated for each pixel, which are passed through a multi-layer perceptron to predict the surface normal for each pixel.}
\label{fig:Skip_Arch}
\end{figure*}


\subsection{Related Work}
The problem of 3D scene understanding has rich history starting from the early works on blocks world~\cite{Roberts65}, to generalized cylinders~\cite{Binford71}, to the work of geons~\cite{Biederman87}. In recent years, most of the work in 3D scene understanding can be divided in two categories: (a) Recovering the 2.5D; (b) Recovering the 3D volumetric objects. The first category of approaches focus on recovering the geometric layout of everyday indoor scenes, e.g., living room, kitchen, bedroom, etc. The goal is to extract a 2.5D representation and extract surface layout~\cite{Hoiem05} or depth of the pixels in the scene. Prior work has sought to recover the overall global shape of the room by fitting a global parametric 3D box~\cite{Hedau2009,Schwing12} or recovering informative edge maps~\cite{Mallya15} that align to the shape of the room, typically based on Manhattan world constraints~\cite{Coughlan00,Kosecka02}. However, such techniques do not recover fine details of object surfaces in the scene.  To recover fine details techniques have sought to output a 2.5D representation (i.e.\ surface normal and depth map) by reasoning about mid-level scene properties, such as discriminative 3D primitives~\cite{Fouhey13a}, convex and concave edges~\cite{fouhey2014unfolding}, and style elements harvested by unsupervised learning~\cite{Fouhey15}. 
Recent approaches have sought to directly predict surface normals and depth via discriminative learning, e.g.,\ with hand-crafted features~\cite{Ladicky14}.  Most similar to our surface normal prediction approach is recent work that trains a CNN to directly predict depth~\cite{Liu15}, jointly predicts surface normals, depth, and object labels~\cite{Eigen15}, or combines CNN features with the global room layout via a predicted 3D box~\cite{Wang15}.

The second category of approaches go beyond  a 2.5D representation and attempt to extract a 3D volumetric representation~\cite{Biederman87,Binford71,Roberts65}. This in line with traditional approaches for object recognition based on 3D model alignment~\cite{Mundy06}.  Parametric models, such as volumetric models~\cite{Lee10}, cuboids~\cite{Xiao12}, joint cuboid and room layout~\cite{Schwing13}, and support surfaces (in RGB-D)~\cite{Guo13} have been proposed. Rendered views of object CAD models over different (textured) backgrounds have been used as training images for CNN-based object detection~\cite{Peng15,Pepik15} and viewpoint estimation~\cite{Su15}. Most similar to us are approaches based on CAD retrieval and alignment.  Approaches using captured RGB-D images from a depth sensor (e.g.\ Kinect) include exemplar detection by rendering depth from CAD and sliding in 3D~\cite{Song14}, 3D model retrieval via exemplar regions matched to object proposals (while optimizing over room layout)~\cite{Guo15}, and training CNNs to predict pose for CAD model alignment~\cite{Gupta15} and to predict object class, location, and pose over rendered CAD scenes~\cite{Papon15}. We address the harder case of alignment to single RGB images.  Recent work include instance detection of a small set of IKEA objects via contour-based alignment~\cite{Lim13}, depth prediction by aligning to renders of 3D shapes via hand-crafted features~\cite{Su14}, object class detection via exemplar matching with mid-level elements~\cite{Aubry14b,Choy15}, and alignment via composition from multiple 3D models using hand-crafted features~\cite{Huang15}. 
More recently CNN-based approaches have developed, such as learning a mapping from CNN features to a 3D light-field embedding space for view-invariant shape retrieval~\cite{Li15} and retrieval using AlexNet~\cite{krizhevsky2012imagenet} pool5 features~\cite{Aubry15}. Also relevant is the approach of Bell and Bala~\cite{Bell15} that trains a Siamese network modeling style similarity to retrieve product images having similar style as a depicted object in an input photo. 

Our work impacts both the categories and bridges the two. First, our skip-network approach (2D $\rightarrow$ 2.5D) uses features from all levels of ConvNet to preserve the fine level details. It provides state of the art performance on surface layout estimation. Our 2.5D$\rightarrow$ 3D approach differs in its development of a CNN that jointly models appearance and predicted surface normals for viewpoint prediction and CAD retrieval.

\subsection{Approach Overview}

Our system takes as input a single 2D image and outputs a set of retrieved object models from a large CAD library matching the style and pose of the depicted objects. 
The system first predicts surface normals capturing the fine details of objects in the scene (Section~\ref{sec:surface_normals}). 
The image, along with the predicted surface normals, are used to retrieve models from the CAD library (Section~\ref{sec:cad_retrieval}). 
We train CNNs for both tasks using the NYU Depth v2~\cite{Silberman12} and rendered views from ModelNet~\cite{Wu15} for the surface normal prediction and CAD retrieval steps, respectively. 
We evaluate both steps and compare against the state of the art in Section~\ref{sec:evaluation}.

\section{Predicting Detailed Surface Normals}
\label{sec:surface_normals}

Our goal is, given a single 2D image $I$, to output a predicted surface normal map $n$ for the image. 
This is a challenging problem due to the large appearance variation of objects, e.g., due to texture, lighting, and viewpoint.

Recently CNN-based approaches have been proposed for this task, achieving state of the art~\cite{Eigen15,Wang15}. 
Wang et al~\cite{Wang15} trained a two-stream network that fuses top-down information about the global room layout with bottom-up information from local image patches. 
While the model recovered the majority of the scene layout, it tended to miss fine details present in the image due to the difficulty of fusing the two streams. 
Eigen and Fergus~\cite{Eigen15} trained a feed-forward coarse-to-fine multi-scale CNN architecture. 
The convolutional layers of the first scale (coarse level) were initialized by training on the object classification task over ImageNet~\cite{Russakovsky15}. 
The remaining network parameters for the mid and fine levels were trained from scratch on the surface normal prediction task using NYU depth~\cite{Silberman12}.
While their approach captured both coarse and fine details, the mid and fine levels of the network were trained on much less data than the coarse level, resulting in inaccurate predictions for many objects.

In light of the above, we seek to better leverage the rich feature representation learned by a CNN trained on large-scale data tasks, such as object classification over ImageNet. 
Recently, Hariharan et al.~\cite{Hariharan15} introduced the hypercolumn representation for the tasks of object detection and segmentation, keypoint localization, and part labeling. 
Hypercolumn feature vectors $h_p(I)$ are formed for each pixel $p$ by concatenating the convolutional responses of a CNN corresponding to pixel location $p$, and capture coarse, mid, and fine-level details. 
Such a representation belongs to the family of skip networks, which have been applied to pixel labeling~\cite{Hariharan15,Long15} and edge detection~\cite{Xie15} tasks. 

We seek to build on the above successes for surface normal prediction. Formally, we seek to learn a function $n_p(I; \theta)$ that predicts surface normals for each pixel location $p$ independently in image $I$ given model parameters $\theta$. 
Given a training set of $N$ image and ground truth surface normal map pairs $\{(I_i, \hat{n}_i)\}_{i=1}^N$, we optimize the following objective: 

\begin{equation}
\min_\theta \sum_{i=1}^N \sum_p || n_p(I_i; \theta) - \hat{n}_{i,p} ||^2.
\end{equation}

We formulate $n_p(I; \theta)$ as a regression network starting from hypercolumn feature $h_p(I)$. Let $c_p^j(I)$ correspond to the outputs of pre-trained CNN layer $j$ at pixel location $p$ given input image $I$. The hypercolumn feature vector is a concatenation of the responses, $h_p(I) = \left(c_p^{j_1}(I), \dots, c_p^{j_\alpha}(I)\right)$ for layers $j_1, \dots, j_\alpha$. 

As shown in Figure~\ref{fig:Skip_Arch}, we train a multi-layer perceptron starting from hypercolumn feature $h_p(I)$ as input. 
Note that the weights of the convolutional layers used to form $h_p(I)$ are updated during training. Also, we normalize the outputs of the last fully-connected layer, which results in minimizing a cosine loss.

Given input vector $x$ and matrix-vector parameters $A_k$ and $b_k$, each layer $k$ produces as output:

\begin{equation}
f_k(x) = ReLU(A_{k}x + b_k),
\end{equation}

\noindent
where element-wise operator $ReLU(z) = max(0,z)$. For our experiments we use three layers in our regression network, setting the output of the last layer as the predicted surface normal $n_p(I; \theta)$. 
Note that Hariharan et al.\ \cite{Hariharan15} learnt weights for a single layer over hypercolumn features. 
We found that having multiple layers captures nonlinearities present in the data and further improves results (c.f.\ Section~\ref{sec:evaluation}).  
Also, note that a fully-convolutional network~\cite{Long15} fuses output class predictions from multiple layers via a directed acyclic graph, whereas we learn regression weights over a concatenation of the layer responses. Our work is similar to Mostajabi et al.~\cite{MostajabiYS15} where they save hypercolumn features to disk and train a multi-layer perceptron. In contrast, ours is an end-to-end pipeline that allows fine tuning of all layers in the network.

\vspace{-0.5cm}
\paragraph{Implementation details and optimization.} 
Given training data, we optimized our network via stochastic gradient descent (SGD) using the publicly-available Caffe source code~\cite{jia2014caffe}. We used a pre-trained VGG-16 network~\cite{SimonyanZ14a} to initialize the weights of our convolutional layers. The VGG-16 network has 13 convolutional layers and 3 fully-connected ({\it fc}) layers. We converted the network to a fully-convolutional one following Long et al.~\cite{Long15}. To avoid confusion with the {\it fc} layers of our multi-layer regression network, we denote {\it fc-6} and {\it fc-7} of VGG-16 as {\it conv-6} and {\it conv-7}, respectively. We used a combination of six different convolutional layers in our hypercolumn feature (we analyze our choices in Section~\ref{sec:evaluation}). 

We constructed mini-batches by resizing training images to $224\times 224$ resolution and randomly sampling pixels from 5 images (1000 pixels were sampled per image). 
We set the starting learning rate to $\epsilon=0.001$ and back propagated through all layers of the network. The random sampling not only ensures that memory remains in bound, but also reduces overfitting due to feature correlation of spatially-neighboring pixels. We employed dropout~\cite{Srivastava14} in the fully-connected layers of the regression network to further reduce overfitting. 
At test time, an image is passed through the network and the output of the last layer are returned as the predicted surface normals.  No further post-processing (outside of ensuring the normals are unit length) is performed on the output surface normals.

\section{Learning Pose and Style for CAD Retrieval}
\label{sec:cad_retrieval}

\begin{figure}[t]
\centering
\includegraphics[width=\linewidth]{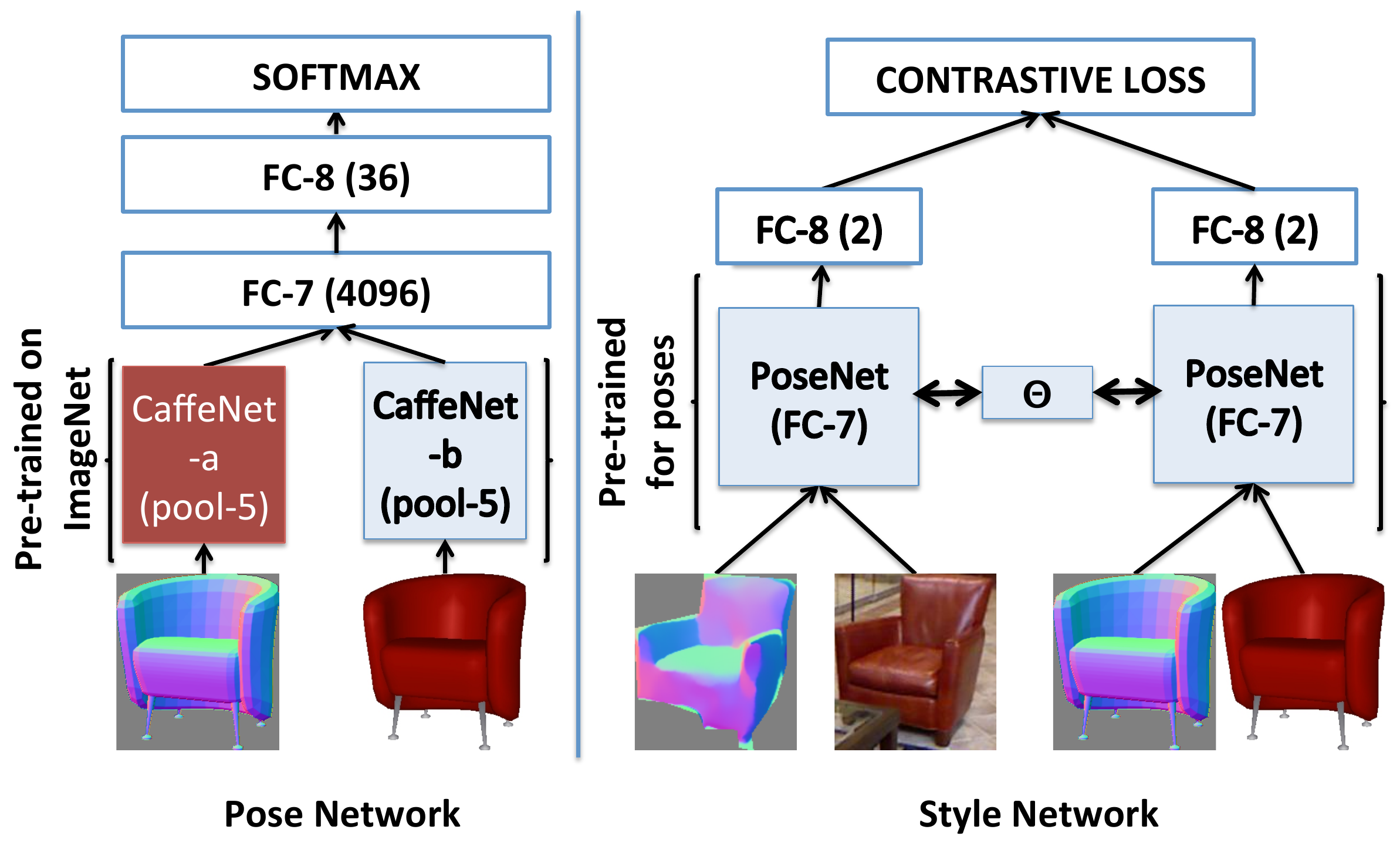}
\caption{
Networks for predicting pose (left) and style (right). 
Our pose network is trained on a set of rendered CAD views and extracted surface normal pairs. 
During prediction, an image and its predicted surface normals are used to predict the object pose. 
For the style network, we train on hand-aligned natural image and CAD rendered view pairs. 
We initialize the style network with the network trained for poses. See text for more details.
}
\label{fig:pose_nw}
\end{figure}

Given a selected image region depicting an object of interest, along with a corresponding predicted surface normal map (Section~\ref{sec:surface_normals}), we seek to retrieve a 3D model from a large object CAD library matching the style and pose of the depicted object. 
This is a hard task given the large number of library models and possible viewpoints of the object. 
While prior work has performed retrieval by matching the image to rendered views of the CAD models~\cite{Aubry14b}, we seek to leverage both the image appearance information and the predicted surface normals.

We first propose a two-stream network to estimate the object pose. This two-stream network takes as input both the image appearance $I$ and predicted surface normals $n(I)$, illustrated in Figure~\ref{fig:pose_nw}(left). Each stream of the two stream network is similar in architecture to CaffeNet~\cite{krizhevsky2012imagenet} upto pool5 layer. We also initialize both the streams using pre-trained ImageNet network.

Note that for surface normals there is no corresponding pre-trained CNN.  Although the CaffeNet model has been trained on images, we have found experimentally (c.f.\ Section~\ref{sec:eval_pose}) that it can also represent well surface normals. As the surface normals are not in the same range as natural images, we found that it is important as a pre-processing step to transform them to be in the expected range. The surface normal values range from $[-1,1]$. We map these scores of surface normals to $[0,255]$ to bring them in same range as natural images. A mean pixel subtraction is done before the image is feed-forward to the network. The mean values for $n_{x}$, $n_{y}$, and $n_{z}$ are computed using the 381 images in train set of NYUD2.

While one could use the pre-trained networks directly for retrieval, such a representation has not been optimized for retrieving CAD models with similar pose and style. We seek to optimize a network to predict pose and style given training data. For learning pose, we leverage the fact that the CAD models are registered to a canonical view so that viewpoint and surface normals are known for rendered views. We generate a training set of sampled rendered views and surface normal maps $\{(I_i, \hat{n}_i)\}_{i=1}^N$ for viewing angles $\{\phi_i\}_{i=1}^N$ for all CAD models in the library. We generate surface normals for each pixel by ray casting to the model faces, which allows us to compute view-based surface normals $\hat{n}$. 

To model pose, we discretize the viewing angles $\phi$ and cast the problem as one of classifying into one of the discrete poses.  We pass the concatenated CaffeNet ``pool5'' features $\bar{c}(I,\hat{n})$ through a sequence of two fully-connected layers, followed by a softmax layer to yield pose predictions $g(I,\hat{n}; \Theta)$ for model parameters $\Theta$.  We optimize a softmax loss over model parameters $\Theta$:

\begin{equation}
\min_{\Theta} -\sum_{i=1}^N \phi_i^T \log(g(I_i,\hat{n}_i; \Theta)).
\end{equation}

Note that during training, we back propagate the loss through all the layers of CaffeNet as well.
Given a trained pose predictor, at test time we pass in image $I$ and predicted surface normals $n(I)$ to yield pose predictions $g(I, n(I); \Theta)$ from the last fully connected layer. 
We can also run our network given RGB-D images, where surface normals are derived from the depth channel. 
We show pose-prediction results for both types of inputs in Section~\ref{sec:eval_pose}. 

Note that a similar network for pose prediction has been proposed for RGB-D input images~\cite{Gupta15}. 
There, they train a network from scratch using normals from CAD for training and query using Kinect-based surface normals during prediction. 
We differ in our use of the pre-trained CaffeNet to represent surface normals and our two-stream network incorporating both surface normal and appearance information. 
We found that due to the differences in appearance of natural images and rendered views of CAD models, simply concatenating the pool5 CaffeNet features hurt performance.  
We augmented the data similar to \cite{Su15} by compositing our rendered views over backgrounds sampled from natural images during training, which improved performance. 

\paragraph{From two-stream pose to siamese style network.} 
While the output of the last fully-connected layer used for pose prediction can be used for retrieval, it has not yet been optimized for style. Inspired by \cite{Bell15}, we seek to model style given a training set of hand-aligned similar and dissimilar CAD model-image pairs. Towards this goal, we extend our two-stream pose network to siamese two-stream network for this task, illustrated in Figure~\ref{fig:pose_nw}(right). Specifically,  let $f$ be the response of the last fully-connected layer of the pose network above. Given similar image-model pairs $(f_p, f_q)$ and dissimilar pairs $(f_q, f_n)$, we seek to optimize the contrastive loss:

\begin{equation}
L(\Theta) = \sum_{(q,p)} L_p(f_q, f_p) + \sum_{(q,n)} L_n(f_q, f_n).
\end{equation}

\noindent
We use the losses $L_p(f_q,f_p)=||f_q-f_p||_2$ and $L_n(f_q,f_n)=\max\left(m-||f_q-f_n||_2,0\right)$, where $m=1$ is a parameter specifying the margin. 
As in \cite{Bell15}, we optimize the above objective via a Siamese network. 
Note that we optimize over pose and style, while \cite{Bell15} optimizes over object class and style for the task of product image retrieval.

For optimization, we apply mini-batch SGD in training using the caffe framework. We followed the standard techniques to train a CaffeNet-like architecture, and back-propagate through all layers. The procedure for training and testing are described in the respective experiment section.

\section{Experiments}
\label{sec:evaluation}

We present an experimental analysis of each component of our pipeline.

\subsection{Surface Normal Estimation} 

The skip-network architecture described in Section~\ref{sec:surface_normals} is used to estimate the surface normals. The VGG-16 network~\cite{SimonyanZ14a} has 13 convolutional layers represented as \{$1_{1}$, $1_{2}$, $2_{1}$, $2_{2}$, $3_{1}$, $3_{2}$, $3_{3}$, $4_{1}$, $4_{2}$, $4_{3}$, $5_{1}$, $5_{2}$, $5_{3}$\}, and three fully-connected layers \{fc-6, fc-7, fc-8\}. As mentioned in Section~\ref{sec:surface_normals}, we convert the pretrained fc-6 and fc-7 layers from VGG-16 to convolutional ones, denoted conv-6 and conv-7, respectively. We use a combination of \{$1_{2}$, $2_{2}$, $3_{3}$, $4_{3}$, $5_{3}$, $7$ \} convolutional layers from VGG-16. We evaluate our approach on NYU Depth v2 dataset~\cite{Silberman12}. There are 795 training images and 654 test images in this dataset. Raw depth videos are also made available by~\cite{Silberman12}. We use the frames extracted from these videos to train our network for the task of surface normal estimation.

For training and testing we use the surface normals computed from the Kinect depth channel by Ladicky et al.~\cite{Ladicky14} over the NYU trainval and test sets.  As their surface normals are not available for the video frames in the training set, we compute normals (from depth data) using the approach of Fouhey et al.~\cite{Fouhey13a}\footnote{Fouhey et al.~\cite{Fouhey13a} used a first-order TGV denoising approach to compute normals from depth data which they used to train their model. We did not use the predicted normals from their approach.}. 

We ignore pixels where depth data is not available during training and testing. 
As shown in \cite{Eigen15,Wang15} data augmentation during training can boost accuracy. 
We performed minimal data augmentation during training. 
We performed left-right flipping of the image and color augmentation, similar to \cite{Wang15}, over the NYU trainval frames only; we did not perform augmentation over the video frames.  
This is much less augmentation than prior approaches~\cite{Eigen15,Wang15}, and we believe we can get additional boost with further augmentation, e.g.\ by employing the suggestions in \cite{Chatfield14}.
Note that the proposed pixel-level optimization also achieves comparable results training on only the 795 images in the training set of the NYUD2 dataset. This is due to the variability provided by pixels in the image as now each pixel act as a data point. 

\begin{figure*}[t]
\centering
\includegraphics[width=\linewidth]{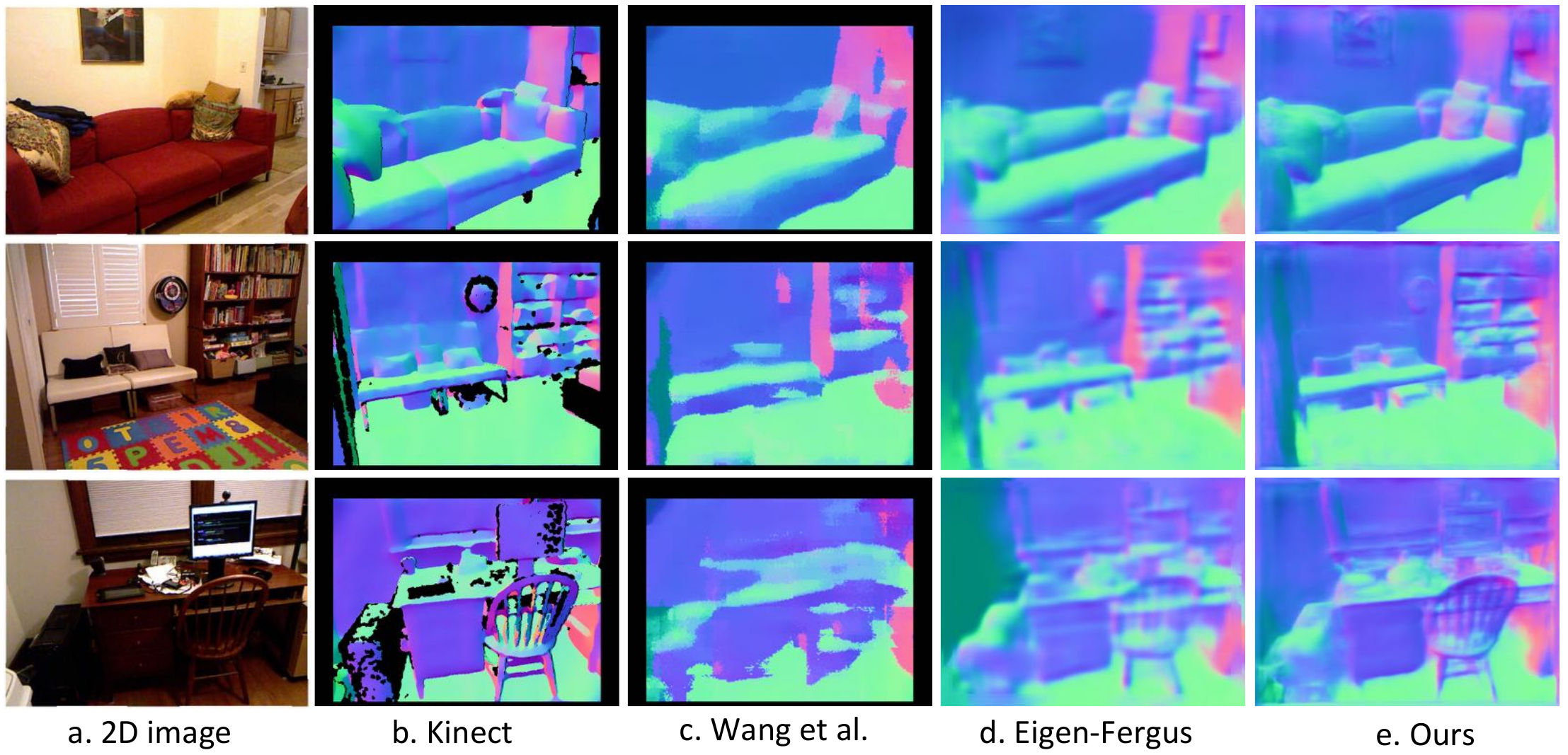}
\caption{Qualitative results for surface normal estimation}
\label{fig:Qual_Surf_Norm}
\vspace{-0.2in}
\end{figure*}

Figure~\ref{fig:Qual_Surf_Norm} shows qualitative results from our approach. Notice that the back of the sofa in row 1 is correctly captured and the fine details of the desk and chair in row 3 are more visible in our approach. 
For quantitative evaluation we use the criteria introduced by Fouhey et al.~\cite{Fouhey13a} to compare our approach against prior work~\cite{Eigen15,Fouhey13a,fouhey2014unfolding,Wang15}. Six statistics are computed over the angular error between the predicted normals and depth-based normals -- \textbf{Mean}, \textbf{Median}, \textbf{RMSE}, \textbf{11.25$^\circ$}, \textbf{22.5$^\circ$}, and \textbf{30$^\circ$} -- using the normals of Ladicky et al.\ as ground truth~\cite{Ladicky14}. The first three criteria capture the mean, median, and RMSE of angular error, where lower is better. The last three criteria capture the percentage of pixels within a given angular error, where higher is better.

In this work, our focus is to capture more detailed surface normal information from the images. We, therefore, not only evaluate our approach on the entire global scene layout as in~\cite{Eigen15,Fouhey13a,fouhey2014unfolding,Wang15}, but we also introduce an evaluation over objects (chair, sofa, and bed) in indoor scene categories. First we show the performance of our approach on the entire global scene layout and compare it with~\cite{Eigen15,Fouhey13a,fouhey2014unfolding,Wang15}. We then compare the surface normals for indoor scene furniture categories (chair, sofa, and bed) against \cite{Eigen15,Wang15}. Finally, we perform an ablative analysis to justify our architecture design choices.
\vspace{-0.1in}

\paragraph{Global Scene Layout:} Table~\ref{tab:nyud2_scene} compares our approach with existing work. We present our results both with and without Manhattan-world rectification to fairly compare against previous approaches, as~\cite{Fouhey13a,fouhey2014unfolding,Wang15} use it and~\cite{Eigen15} do not. Similar to~\cite{Fouhey13a}, we rectify our normals using the vanishing point estimates from Hedau et al.~\cite{Hedau2009}. Interestingly, our approach performs worse with Manhattan-world rectification (unlike Fouhey et al.\ \cite{Fouhey13a}). 
Our network architecture predicts room layout automatically, and appears to be better than using vanishing point estimates. Though capturing scene layout was not our objective, our work out-performs previous approaches on all evaluation criteria. 

\begin{table}
\small{
\setlength{\tabcolsep}{3pt}
\def\arraystretch{1.2}
\center
\begin{tabular}{@{}l c c c c c c }
\toprule
\textbf{NYUDv2 test}  & Mean  &   Median & RMSE &  11.25$^\circ$ & 22.5$^\circ$ &  30$^\circ$ \\
\midrule
Eigen-Fergus~\cite{Eigen15} 	      & 23.7      &	15.5	 &    -	 &   39.2     & 62.0	     &	71.1	\\ 
Fouhey et al.~\cite{Fouhey13a}	      &  35.3     &	31.2	 &   41.4	 &   16.4      &	  36.6   &	48.2	\\ 
\midrule
Ours		      &	  \textbf{19.8}     &	\textbf{12.0}	 &   \textbf{28.2}	 &   \textbf{47.9}     &	   \textbf{70.0}  & \textbf{77.8}		\\  	

\midrule
\midrule
\textbf{Manhattan World} \\ 
Wang et al.~\cite{Wang15}	      &   26.9    &	14.8	 &   - &   42.0     &	 61.2    &	68.2	\\ 
Fouhey et al.~\cite{fouhey2014unfolding}	      & 35.2      &	17.9	 &   49.6	 &    40.5    &	 54.1    &	58.9	\\ 
Fouhey et al.~\cite{Fouhey13a}         &   36.3    &	19.2	 &   50.4	 &  39.2       &	52.9     &	57.8	\\ 
\midrule
Ours	      &  \textbf{23.9}     &	\textbf{11.9}	 &  \textbf{35.9}	 &  \textbf{48.4}      &	 \textbf{66.0}    &	\textbf{72.7}	\\ 
\bottomrule
\end{tabular}
\vspace{3pt}
\caption{
NYUv2 surface normal prediction: Global scene layout.
Note that the results of Eigen-Fergus~\cite{Eigen15} were taken from an earlier arXiv version of their paper. In their most recent version, they achieved improved results using a VGG-16 network in their architecture. We out-perform their latest results by 1-3\% on all evaluation criteria, and will update their results in our final paper version.}
\label{tab:nyud2_scene}
}
\vspace{-0.1cm}
\end{table}
\begin{table}
\small{
\setlength{\tabcolsep}{3pt}
\def\arraystretch{1.2}
\center
\begin{tabular}{@{}l c c  c  c c c }
\toprule
\textbf{NYUDv2 test}  & Mean  &   Median &  RMSE  & 11.25$^\circ$ & 22.5$^\circ$ &  30$^\circ$ \\
\midrule
\textbf{Chair}\\
Wang et al.~\cite{Wang15}  	      &   44.7    &	35.8   &    54.9	 &	 14.2        &	 34.3    &	44.3	\\ 
Eigen-Fergus~\cite{Eigen15}	      &   38.2    &	32.5   &    46.3         &	  14.4         &	34.9     &	46.6	\\ 
Ours		      &  \textbf{32.0}     &	\textbf{24.1}    &    \textbf{40.6}	 &  \textbf{21.2}	         &	\textbf{47.3}     &	\textbf{58.5}	\\ 
\midrule
\midrule
\textbf{Sofa}\\
Wang et al.~\cite{Wang15}  	      &  36.0     &	27.6   &   45.4	 &	21.6         &	 42.6    &	53.1	\\ 
Eigen-Fergus~\cite{Eigen15}	      &   27.0    &	21.3   &	34.0 &	 25.5        &	 52.4    &	63.4	\\ 
Ours		      &   \textbf{20.9}    &	\textbf{15.9}    &	\textbf{27.0} &	    \textbf{34.8}     &	\textbf{66.1}     &	\textbf{77.7}	\\ 
\midrule
\midrule
\textbf{Bed}\\
Wang et al.~\cite{Wang15}  	      &  28.6     &	18.5   &   38.7	 &	 34.0       &	56.4     &	65.3	\\ 
Eigen-Fergus~\cite{Eigen15}	      &   23.1    &	16.3   &   30.8	 &	  36.4        &	 62.0    &	72.6	\\ 
Ours		      &  \textbf{19.6}     &	\textbf{13.4}    &   \textbf{26.9}	 &	\textbf{43.5}         &	\textbf{69.3}     &	\textbf{79.3}	\\ 
\bottomrule
\end{tabular}
\vspace{3pt}
\caption{
NYUv2 surface normal prediction: Local object layout.
}
\label{tab:nyud2_obj}
}
\vspace{-0.1cm}
\end{table}
\vspace{-0.1in}
\paragraph{Local Object Layout:} The existing surface normal literature is focussed towards the scene layout. In this work, we stress the importance of fine details in the scene generally available around objects. We, therefore, evaluated the performance of our approach in the object regions by considering only those pixels which belong to a particular object. Here we show the performance on chair, sofa and bed. 
Table~\ref{tab:nyud2_obj} shows comparison of our approach with Wang et al.~\cite{Wang15} and Eigen and Fergus~\cite{Eigen15}. We achieve performance around \textbf{3-10\%} better than previous approaches on all statistics for all the objects.

\paragraph{Ablative Analysis:} We analyze how different sets of convolutional layers influence the performance of our approach. Table~\ref{tab:nyud2_pixel} shows some of our analysis. We chose a combination of layers from low, mid, and high parts of the VGG network. Clearly from the experiments, we need a combination of different low, mid, high layers to capture rich information present in the image. 

\begin{table}
\footnotesize{
\setlength{\tabcolsep}{3pt}
\def\arraystretch{1.2}
\center
\begin{tabular}{@{}l c c c c c c }
\toprule
\textbf{NYUDv2 test}  & Mean  &   Median & RMSE &  11.25$^\circ$ & 22.5$^\circ$ &  30$^\circ$ \\
\midrule
\{$1_{1}$, $1_{2}$\} 	      & 44.4    &	42.7	 &  49.3	 & 4.1     &	16.5   &	28.2	\\ 
\{$1_1$, $1_2$, $3_3$\}      & 30.2   &	24.7	 & 37.7 	 & 23.1     &	 46.2  &	58.4	\\ 
\{$1_1$, $1_2$, $5_3$\} 	      &  22.6   &	15.3	 &  30.5	 &  39.1    &	  63.4 &	73.1	\\ 
\{$1_1$, $1_2$, $3_3$, $5_3$\}	      & 21.3    &	13.9	 &  29.2	 &   42.3   &	67.0   &		76.0\\ 
\{$1_2$, $3_3$, $5_3$\}	    &  21.3    &	14.0	 &   29.3	 &   42.0     &	   66.7  & 75.8		\\ 
\{$1_2$, $2_2$, $3_3$, $4_3$, $5_3$\}  	      &  20.9     &  13.6 &   28.0	 &   43.1    &	   67.9  & 77.0		\\ 
\midrule
\{$1_2$, $2_2$, $3_3$, $4_3$, $5_3$, $7$\}  	      &	  \textbf{19.8}     &	\textbf{12.0}	 &   \textbf{28.2}	 &   \textbf{47.9}     &	   \textbf{70.0}  & \textbf{77.8}		\\  	

\bottomrule
\end{tabular}
\vspace{3pt}
\caption{
NYUv2 surface normal prediction: Ablative Analysis.
}
\label{tab:nyud2_pixel}
}
\vspace{-0.1cm}
\end{table}

\subsection{Pose Estimation} 
\label{sec:eval_pose}
We evaluated the approach described in Section~\ref{sec:cad_retrieval} to estimate the pose of a given object. 
We trained the pose network using CAD models from Princeton ModelNet~\cite{Wu15} as training data, and used 1260 models for chair, 526 for sofa, and 196 for bed. 
For each model, we rendered 144 different views corresponding to 4 elevation and 36 azimuth angles. 
We designed the network to predict one of the 36 azimuth angles, which we treated as a 36-class classification problem. 
Note that we trained separate pose networks for the chair, sofa, and bed classes. At test time, we forward propagated the selected region from the image, along with its predicted surface normals, and selected the angle with maximum prediction score.  We evaluated our approach using the annotations from Guo and Hoiem~\cite{Guo13} where they manually annotated the NYUD2 dataset with aligned 3D CAD models for the categories of interest. 

\begin{figure}[h]
\centering
\includegraphics[width=\linewidth]{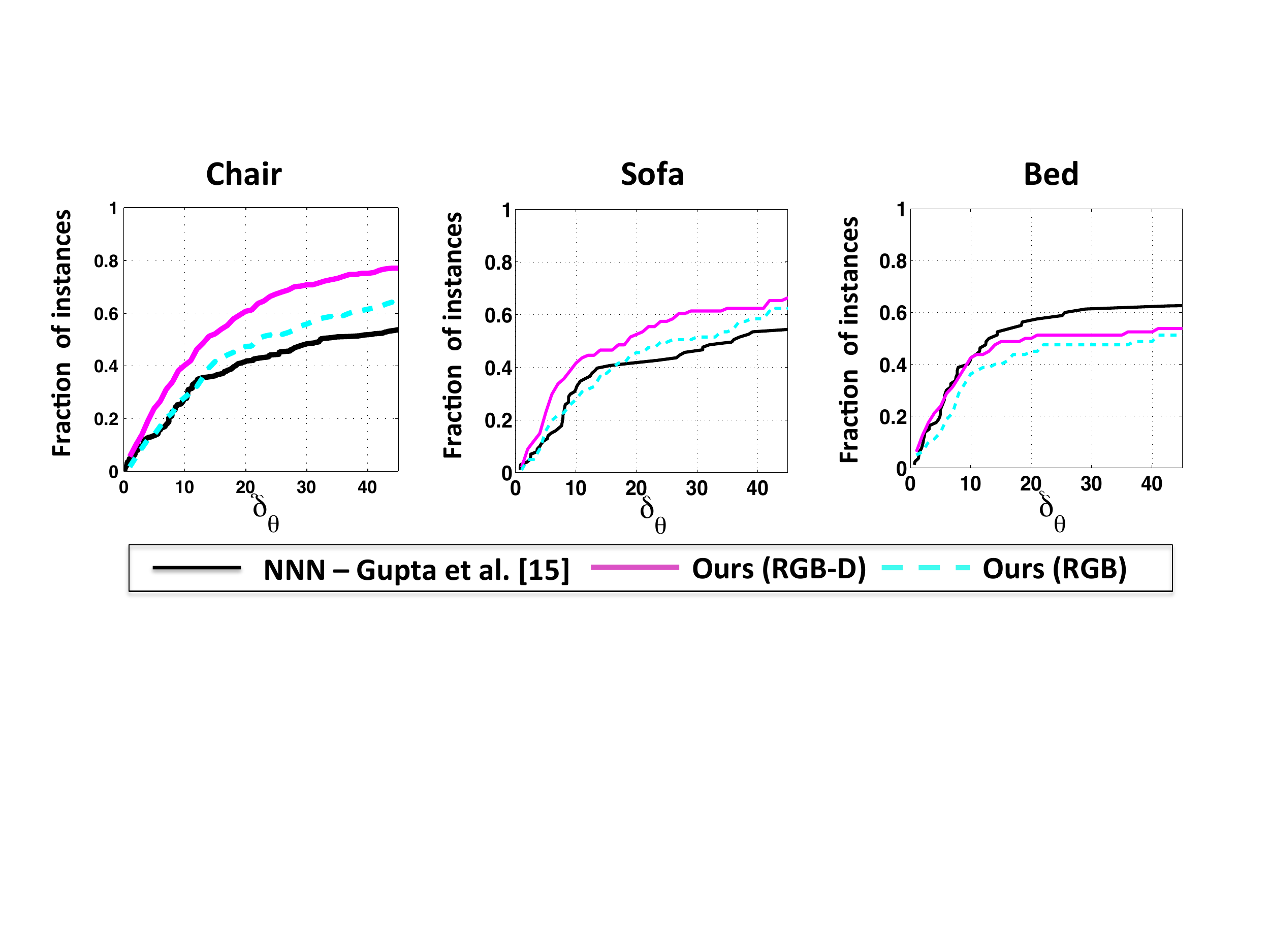}
\caption{
Pose prediction on val set. 
We plot the fraction of instances with predicted pose angular error less than $\delta_\theta$ as a function of $\delta_\theta$. Similar to~\cite{Gupta15} we consider only those objects which have valid depth pixels for more than 50\%.
}
\label{fig:pose_cad}
\end{figure}
  
\paragraph{Comparison on NYUD2 val set:}  Figure~\ref{fig:pose_cad} shows a quantitative evaluation of our approach on the NYUD2 val set. Using the criteria introduced in Gupta et al~\cite{Gupta15}, we plot the fraction of instances with predicted pose angular error less than $\delta_\theta$ as a function of $\delta_\theta$ (higher is better). 
We compare our approach with Gupta et al~\cite{Gupta15} who showed results of pose estimation on the NYUD2 val set for objects with at least 50\% valid depth pixels. 
Note that we trained our skip-network for surface normals using the 381 images of the NYUD2 train set. 
We clearly out-perform the baseline using RGB-only and RGB-D for chairs and sofas.

\begin{figure}[h]
\centering
\includegraphics[width=\linewidth]{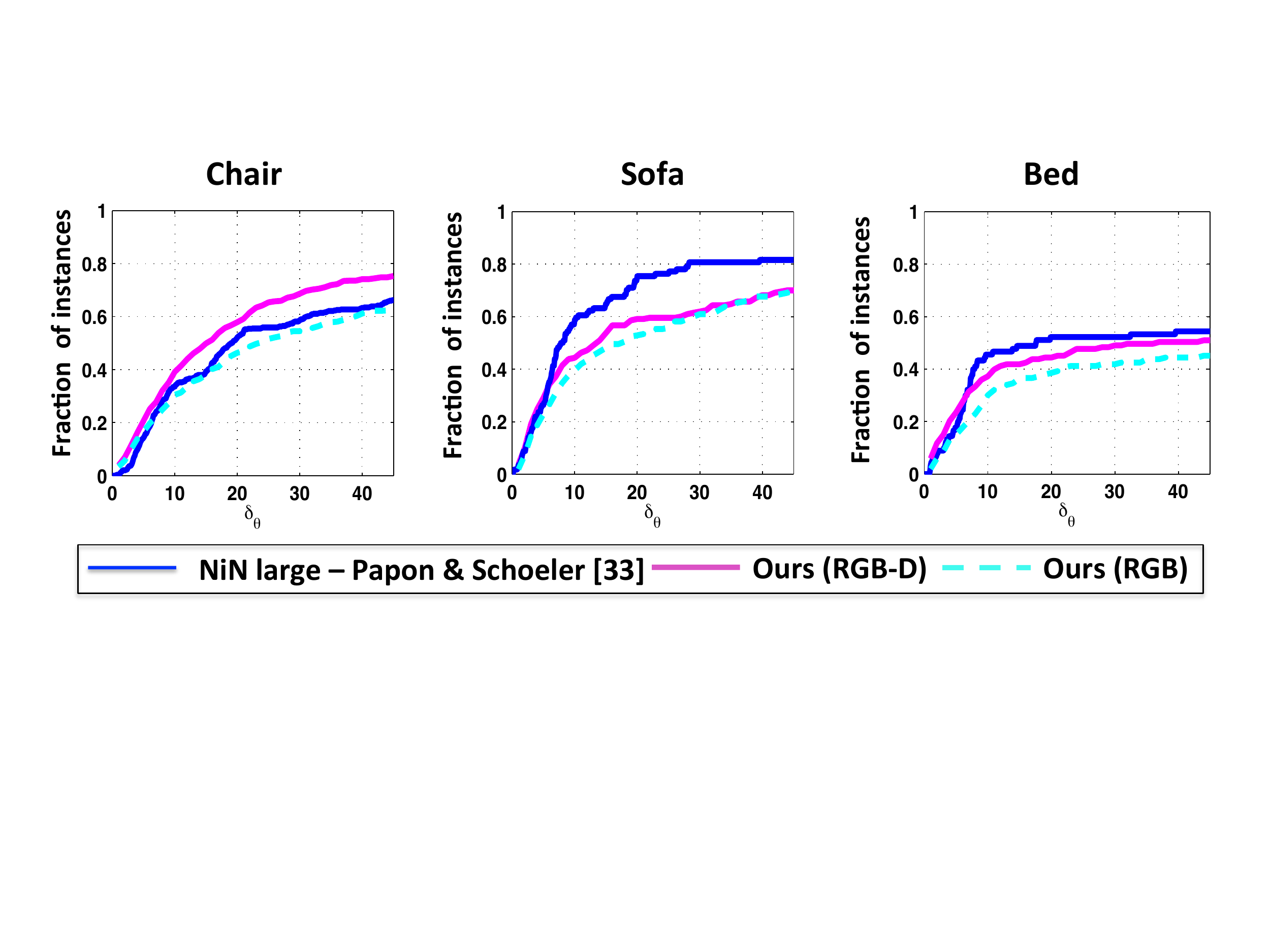}
\caption{
Pose prediction on test set. 
We plot the fraction of instances with predicted pose angular error less than $\delta_\theta$ as a function of $\delta_\theta$. Similar to~\cite{Gupta15} we consider only those objects which have valid depth pixels for more than 50\%.
}
\label{pose_est_test}
\end{figure}

\paragraph{Comparison on NYUD2 test set:} 
Unfortunately, we cannot directly compare the approaches of~\cite{Gupta15} and~\cite{Papon15} for pose estimation. While Gupta et al.~\cite{Gupta15} reported performance on the NYUD2 val set, Papon and Schoeler~\cite{Papon15} reported performance on the test set. 
We evaluated our approach on the val and test sets separately to directly compare against both methods. Figure~\ref{pose_est_test} shows the comparison of our approach against Papon and Schoeler~\cite{Papon15} on the test set (we trained using the NYUD2 trainval set) and shows that our approach is competitive for both RGB-D and RGB. 

\begin{figure}[h]
\centering
\includegraphics[width=\linewidth]{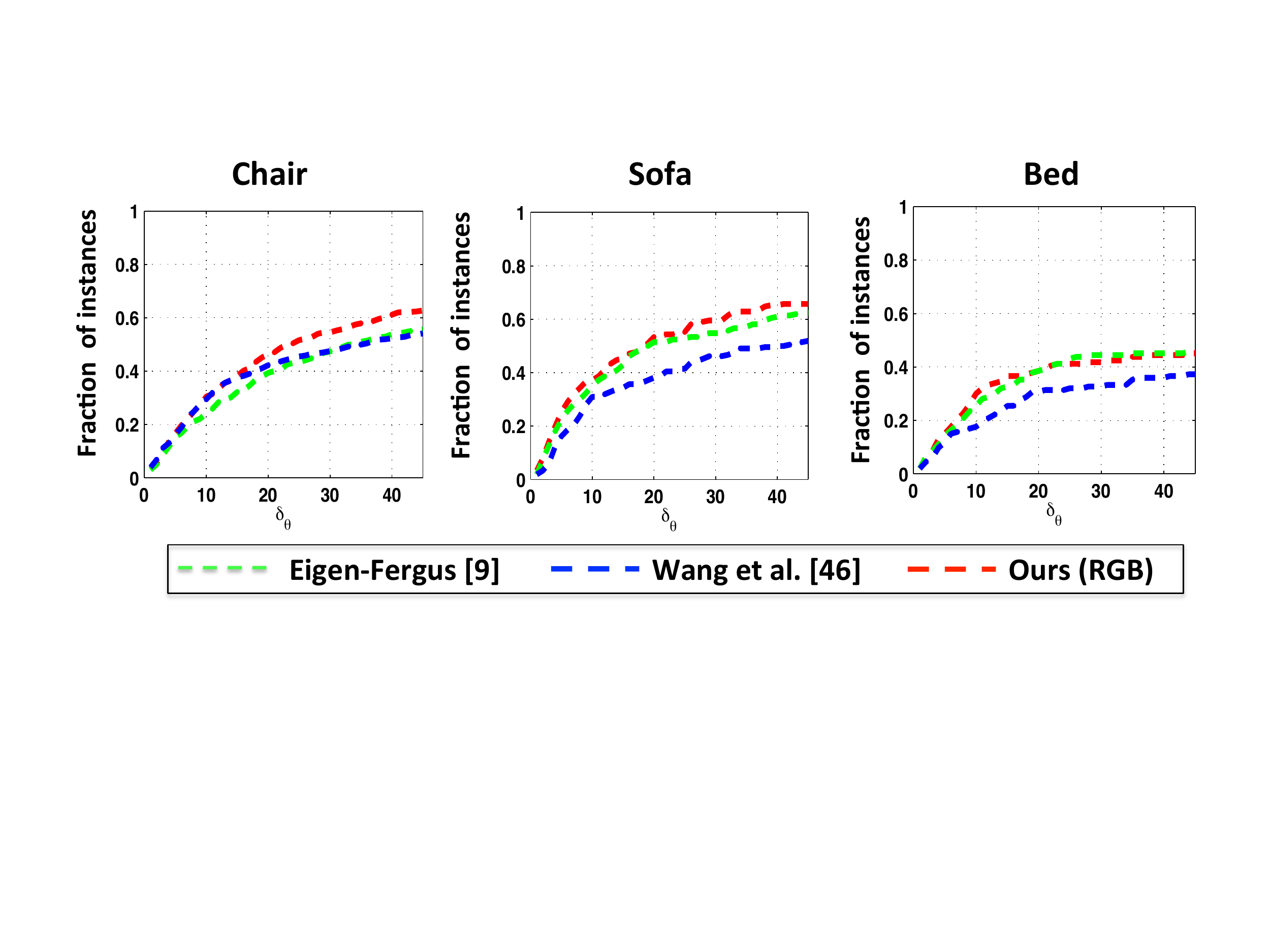}
\caption{
Pose prediction for different surface normal predictions. We plot the fraction of instances with predicted pose angular error less than $\delta_\theta$ as a function of $\delta_\theta$. Similar to~\cite{Gupta15} we consider only those objects which have valid depth pixels for more than 50\%.
}
\label{pose_est_norm}
\vspace{-0.2in}
\end{figure}

\paragraph{Varying predicted surface normals:} 
We analyze how different surface normal prediction algorithms affect the accuracy of predicting object pose.  
Since no real-world data was used for training our pose estimation network (we only used CAD model rendered views), we can perform this experiment without any bias with respect to the surface normal prediction algorithm. 
Figure~\ref{pose_est_norm} shows a comparison of our approach, along with Eigen and Fergus~\cite{Eigen15} and Wang et al.~\cite{Wang15}. 
Notice that better surface normal prediction results in better object pose predictions. 

\begin{figure}[h]
\centering
\includegraphics[width=\linewidth]{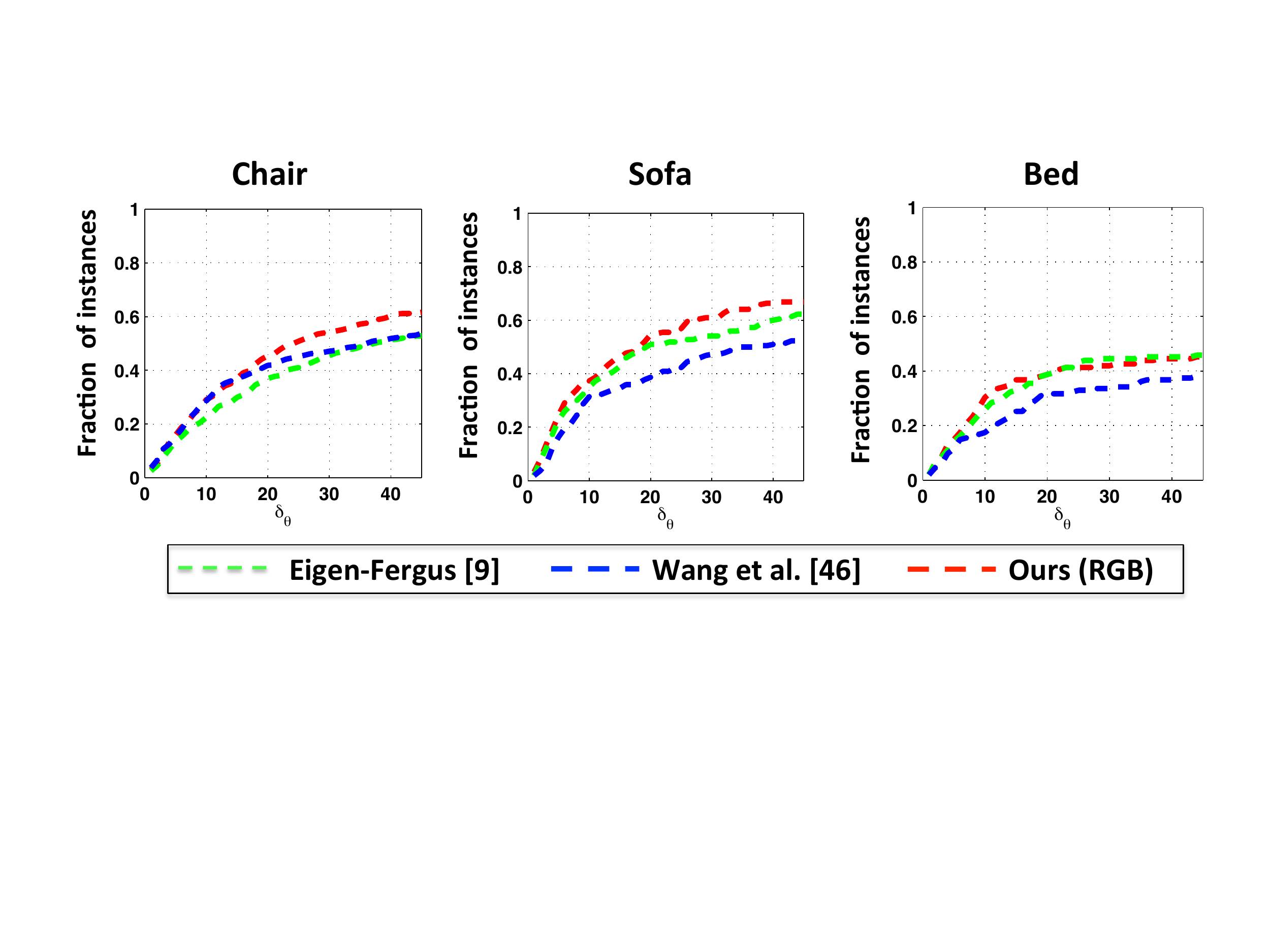}
\caption{
Pose prediction for all images irrespective of depth constraint. We plot the fraction of instances with predicted pose angular error less than $\delta_\theta$ as a function of $\delta_\theta$. In this analysis, we consider all the objects irrespective of valid depth data.
}
\label{pose_est_no_dep_norm}
\vspace{-0.2in}
\end{figure}

\paragraph{Removing depth constraint in evaluation:} 
So far we have ignored test examples having less than 50\% valid depth pixels since prior approaches based on RGB-D data require valid depth for object pose prediction. 
The predicted normals gives us an added advantage to consider all examples irrespective of available depth information. 
In this experiment we compare our approach for pose estimation without any depth-based criteria. 
Figure~\ref{pose_est_no_dep_norm} shows the performance of different surface normal approaches on NYUD2 test set for {\bf all test examples}.

\begin{figure}[t]
\centering
\includegraphics[width=\linewidth]{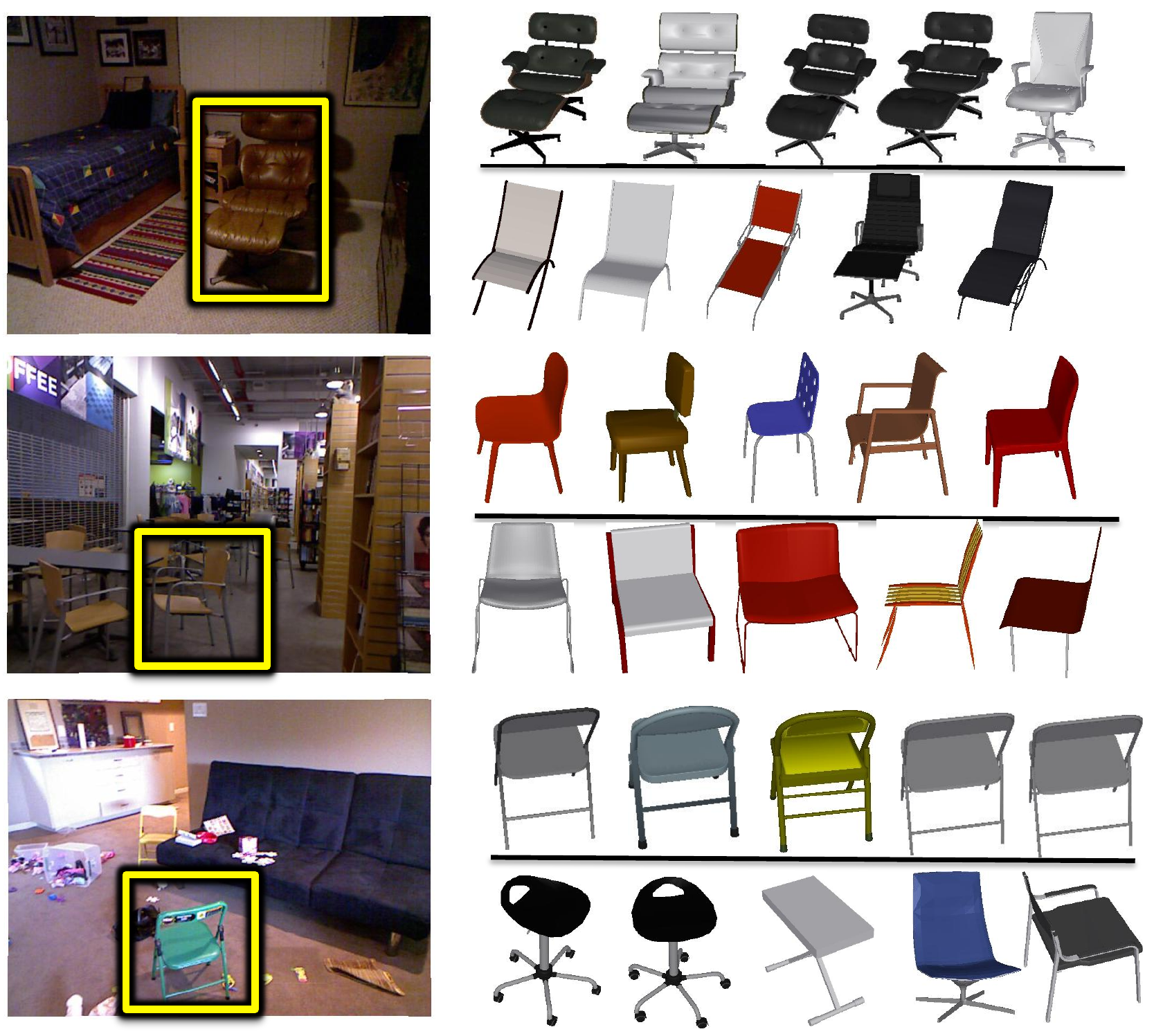}
\caption{For each example, the top row shows CAD models retrieved using fc-7 of Pose Network and the bottom row shows the result of nearest-neighbor retrieval using predicted surface normals.}
\label{fig:pose_retrieval}
\vspace{-0.2in}
\end{figure}

\paragraph{Comparison with nearest neighbors: } 
We compare our approach with nearest neighbors using surface normals and CaffeNet pool-5 features to retrieve a 3D model that is similar in pose and style. When using surface normals, we slide the CAD models on the given bounding box of the image to determine the correct location and scale for the model. The CAD model rendered views are resized such that the maximum dimension is 40 pixels. We tried two approaches for scoring the windows for the sliding-window approach - 1) compute dot product; 2) compute the angular error between the two, and then compute the percentage of pixels within $30^\circ$ angular error (we call this criteria `Geom'). To penalize smaller windows we compute the product of the scores and overlap (IoU) between the window and original box. This ensures there is no bias towards smaller windows in the sliding-window approach. Finally, we prune similar CAD models within a $20^\circ$ azimuthal angle. 

To capture appearance information during retrieval, we used the CaffeNet pool-5 features for both CAD models and the given bounding box. We computed the cosine distance between pool-5 features. Both scores from appearance and surface normals were combined into a final score by weighted averaging.

We evaluated on the chairs in the test set. In this experiment, we only considered chairs having more than 50\% of valid depth pixels. We report area under the pose prediction curve in Table~\ref{tab:pose_comp_nn_test}. 
Notice that the `Geom' criteria outperforms dot product. 
Also, combining appearance information boosts the performance of predicted normals but hurts the performance of normals from depth. Note that our \textbf{PoseNet} trained on just \textbf{RGB} is comparable to the results obtained using nearest neighbors for RGB-D. 

\begin{table}[h]
\small{
\setlength{\tabcolsep}{3pt}
\def\arraystretch{1.2}
\center
\begin{tabular}{@{}l c c c c c}
\toprule
\textbf{NYUD2 test}	  &  Dot-Product		& Dot-Product      &   Geom   	& 	Geom  \\
			  &   (1 NN)			& (K-NN)	   &  (1 NN)	& 	(K-NN) \\
\midrule	
Random			  &	0.13			&	0.13	   &	0.13	&	0.13	     \\
Normals (ours)		  &	0.21			&	0.22	   &	0.31	&	0.30	     \\
Normals (depth)		  &	0.33			&	0.31	   &	0.41	&	0.44	     \\
\midrule
App. (pool-5)	  	  &	0.26			&	0.28	   &	0.26 	&	0.28	     \\
\midrule
ours+pool-5	  	  &	0.25			&	0.28	   &	0.29	&	0.32	     \\
depth+pool-5	  	  &	0.30			&	0.33	   &	0.36	&	0.38	     \\
\bottomrule
\end{tabular}
\vspace{3pt}
\caption{Area under the fraction of instances versus angular error $\delta_\theta$ curve. Similar to~\cite{Gupta15,Papon15}, we consider only those objects which have valid depth pixels for more than 50\%.   For K-NN we used K = 35. Note that for App.\ (pool-5), we did not use the `Geom' criteria but copied the scores of dot product in it. Our PoseNet (RGB) and (RGB-D) achieves 0.43 and 0.55 respectively, which is higher than the proposed nearest neighbor approaches.}
\label{tab:pose_comp_nn_test}
}
\end{table}

\subsection{Style Estimation} 
\begin{figure}[t]
\centering
\includegraphics[width=\linewidth]{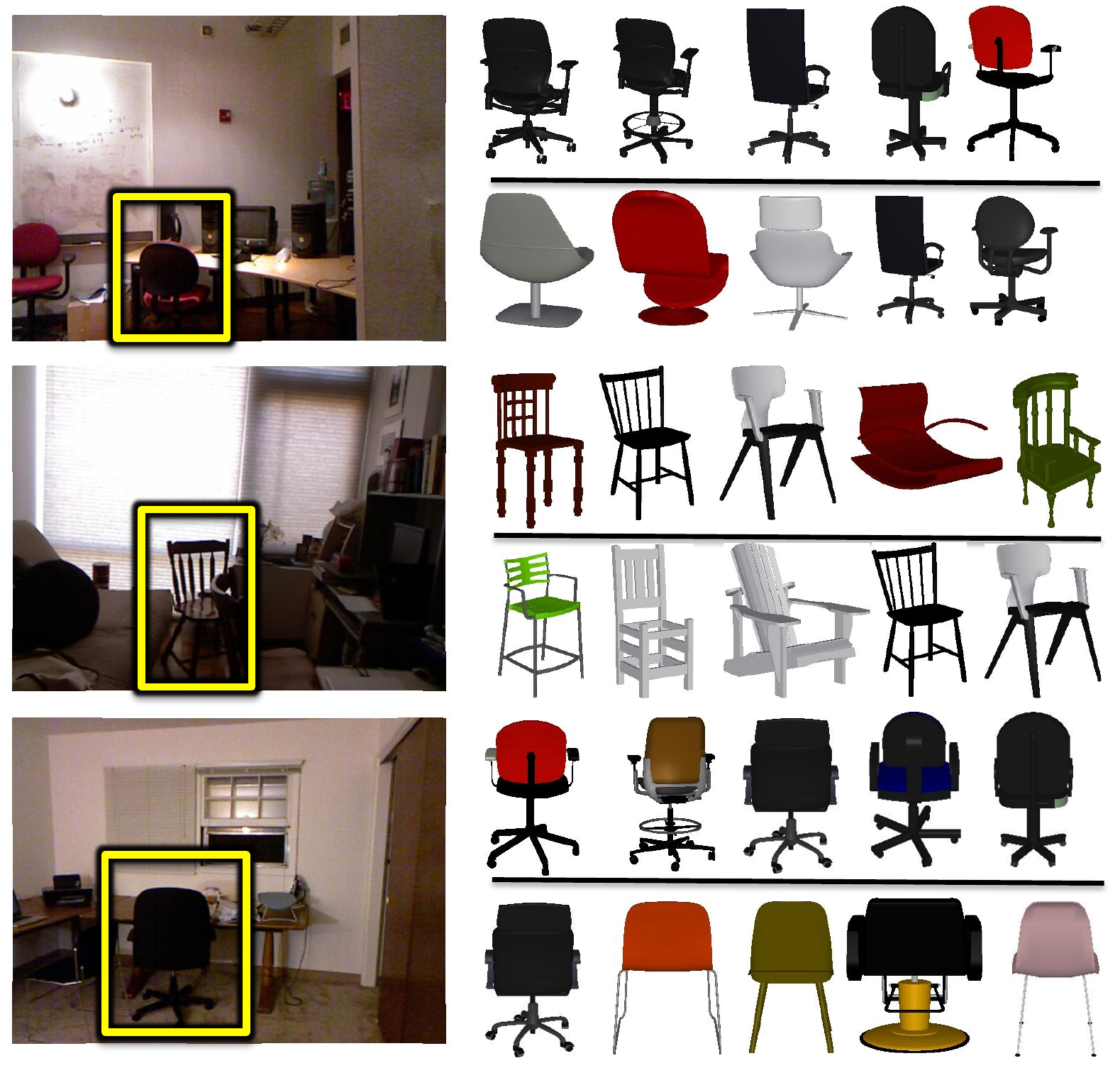}
\caption{Style re-ranking. For each example the top row shows the top-5 CAD models obtained using our Style Network and the bottom row shows the original retrievals using the Pose Network.}
\label{fig:style_rerank}
\vspace{-0.1cm}
\end{figure}

We used the style network described in Section~\ref{sec:cad_retrieval} to determine the style of objects. To reduce the search space, we use this network to re-rank the top-$N$ output of the CAD models retrieved using the fc-7 feature of the pose network. We evaluate our style network using chairs as chairs span a large variety of styles~\cite{Aubry14b}. To train the model we hand-labeled images in the NYUD2 training set with models having very similar style. To assist with the labeling we used our pose network to retrieve similar CAD models over the NYU training set. For each example we looked at the top-30 retrieved CAD models and manually labeled if a particular CAD model is similar to the input example or not. We used these labels to train our style network using the contrastive loss. Figure~\ref{fig:style_rerank} shows qualitative examples of our re-ranking via the style network. Note that the network is able to boost the ranking of similar examples, e.g., the chairs having wheels in the first and last row have different styles in the initial retrieved examples of the pose network. With the re-ranking, we are able to see chairs with wheels consistently.

\section{Conclusion}

We have demonstrated a successful feed-forward approach for 3D object recognition in 2D images via 2.5D surface normal prediction. 
Our skip-network approach for surface normal prediction recovers fine object detail and achieves state of the art on the challenging NYU depth benchmark. 
We formulated a two-stream pose network that jointly reasons over the 2D image and predicted surface normals, and achieves pose prediction accuracy that is comparable to existing approaches based on RGB-D images. 
When we apply our pose network to RGB-D image data, we surpass the state of the art for the pose prediction task.
Finally, our pose-style network shows promising results in retrieving CAD models matching both the depicted object style and pose. 
Our accurate surface normal predictions open up the possibility of having reliable 2.5D predictions for most natural images, which may have impact on applications in computer graphics and, ultimately, for the goal of full 3D scene understanding.

\vspace{0.1cm}
\noindent\textbf{Acknowledgements:} This work was partially supported by grants from NSF IIS-1320083 and ONR MURI N000141612007. We thank Saining Xie for discussion on skip-network architectures, David Fouhey for providing code to compute normals from Kinect data, and Saurabh Gupta for help with the pose estimation evaluation setup.

{\small
\bibliographystyle{ieee}
\bibliography{shortstrings,references}

\begin{thebibliography}{10}\itemsep=-1pt

\bibitem{Aubry14b}
M.~Aubry, D.~Maturana, A.~A. Efros, B.~C. Russell, and J.~Sivic.
\newblock Seeing {3D} chairs: Exemplar part-based {2D-3D} alignment using a
  large dataset of {CAD} models.
\newblock In {\em CVPR}, 2014.

\bibitem{Aubry15}
M.~Aubry and B.~C. Russell.
\newblock Understanding deep features with computer-generated imagery.
\newblock In {\em {ICCV}}, 2015.

\bibitem{Bell15}
S.~Bell and K.~Bala.
\newblock Learning visual similarity for product design with convolutional
  neural networks.
\newblock {\em ACM Transactions on Graphics (Proceeding of SIGGRAPH)}, 2015.

\bibitem{Biederman87}
I.~Biederman.
\newblock Recognition by components: a theory of human image interpretation.
\newblock {\em Pyschological review}, 94:115--147, 1987.

\bibitem{Binford71}
T.~O. Binford.
\newblock Visual perception by computer.
\newblock In {\em {IEEE conference on Systems and Control}}, 1971.

\bibitem{Chatfield14}
K.~Chatfield, K.~Simonyan, A.~Vedaldi, and A.~Zisserman.
\newblock Return of the devil in the details: Delving deep into convolutional
  nets.
\newblock In {\em British Machine Vision Conference}, 2014.

\bibitem{Choy15}
C.~B. Choy, M.~Stark, S.~Corbett-Davies, and S.~Savarese.
\newblock Object detection with {2D-3D} registration and continuous viewpoint
  estimation.
\newblock In {\em {CVPR}}, 2015.

\bibitem{Coughlan00}
J.~Coughlan and A.~Yuille.
\newblock The {Manhattan} world assumption: Regularities in scene statistics
  which enable {Bayesian} inference.
\newblock In {\em {NIPS}}, 2000.

\bibitem{Eigen15}
D.~Eigen and R.~Fergus.
\newblock Predicting depth, surface normals and semantic labels with a common
  multi-scale convolutional architecture.
\newblock In {\em ICCV}, 2015.

\bibitem{Fouhey13a}
D.~F. Fouhey, A.~Gupta, and M.~Hebert.
\newblock Data-driven {3D} primitives for single image understanding.
\newblock In {\em ICCV}, 2013.

\bibitem{fouhey2014unfolding}
D.~F. Fouhey, A.~Gupta, and M.~Hebert.
\newblock Unfolding an indoor origami world.
\newblock In {\em ECCV}, 2014.

\bibitem{Fouhey15}
D.~F. Fouhey, A.~Gupta, and M.~Hebert.
\newblock Single image {3D} without a single {3D} image.
\newblock In {\em {ICCV}}, 2015.

\bibitem{Guo13}
R.~Guo and D.~Hoiem.
\newblock Support surface prediction in indoor scenes.
\newblock In {\em ICCV}, 2013.

\bibitem{Guo15}
R.~Guo, C.~Zou, and D.~Hoiem.
\newblock Predicting complete 3d models of indoor scenes.
\newblock In {\em arXiv:1504.02437}, 2015.

\bibitem{Gupta15}
S.~Gupta, P.~A. Arbel{\'{a}}ez, R.~B. Girshick, and J.~Malik.
\newblock Aligning {3D} models to {RGB-D} images of cluttered scenes.
\newblock In {\em CVPR}, 2015.

\bibitem{Hariharan15}
B.~Hariharan, P.~Arbel{\'{a}}ez, R.~Girshick, and J.~Malik.
\newblock Hypercolumns for object segmentation and fine-grained localization.
\newblock In {\em CVPR}, 2015.

\bibitem{Hedau2009}
V.~Hedau, D.~Hoiem, and D.~Forsyth.
\newblock Recovering the spatial layout of cluttered rooms.
\newblock In {\em ICCV}, 2009.

\bibitem{Hoiem05}
D.~Hoiem, A.~Efros, and M.~Hebert.
\newblock Automatic photo pop-up.
\newblock In {\em {SIGGRAPH}}, 2005.

\bibitem{Huang15}
Q.~Huang, H.~Wang, and V.~Koltun.
\newblock Single-view reconstruction via joint analysis of image and shape
  collections.
\newblock {\em ACM Transactions on Graphics (Proceeding of SIGGRAPH)}, 34(4),
  2015.

\bibitem{jia2014caffe}
Y.~Jia, E.~Shelhamer, J.~Donahue, S.~Karayev, J.~Long, R.~Girshick,
  S.~Guadarrama, and T.~Darrell.
\newblock Caffe: Convolutional architecture for fast feature embedding.
\newblock {\em arXiv preprint arXiv:1408.5093}, 2014.

\bibitem{Kosecka02}
J.~Kosecka and W.~Zhang.
\newblock Video compass.
\newblock In {\em {ECCV}}, 2002.

\bibitem{krizhevsky2012imagenet}
A.~Krizhevsky, I.~Sutskever, and G.~E. Hinton.
\newblock Imagenet classification with deep convolutional neural networks.
\newblock In {\em NIPS}, 2012.

\bibitem{Ladicky14}
L.~Ladicky, B.~Zeisl, and M.~Pollefeys.
\newblock Discriminatively trained dense surface normal estimation.
\newblock In {\em {ECCV}}, 2014.

\bibitem{Lee10}
D.~C. Lee, A.~Gupta, M.~Hebert, and T.~Kanade.
\newblock Estimating spatial layout of rooms using volumetric reasoning about
  objects and surfaces.
\newblock In {\em {NIPS}}, 2010.

\bibitem{Li15}
Y.~Li, H.~Su, C.~R. Qi, N.~Fish, D.~Cohen-Or, and L.~J. Guibas.
\newblock Joint embeddings of shapes and images via {CNN} image purification.
\newblock {\em ACM Transactions on Graphics (Proceeding of SIGGRAPH Asia)},
  2015.

\bibitem{Lim13}
J.~J. Lim, H.~Pirsiavash, and A.~Torralba.
\newblock Parsing {IKEA} objects: Fine pose estimation.
\newblock In {\em ICCV}, 2013.

\bibitem{Liu15}
F.~Liu, C.~Shen, and G.~Lin.
\newblock Deep convolutional neural fields for depth estimation from a single
  image.
\newblock In {\em CVPR}, 2015.

\bibitem{Long15}
J.~Long, E.~Shelhamer, and T.~Darrell.
\newblock Fully convolutional models for semantic segmentation.
\newblock In {\em CVPR}, 2015.

\bibitem{Mallya15}
A.~Mallya and S.~Lazebnik.
\newblock Learning informative edge maps for indoor scene layout prediction.
\newblock In {\em {ICCV}}, 2015.

\bibitem{Marr82}
D.~Marr.
\newblock {\em Vision}.
\newblock W. H. Freeman and Company, 1982.

\bibitem{MostajabiYS15}
M.~Mostajabi, P.~Yadollahpour, and G.~Shakhnarovich.
\newblock Feedforward semantic segmentation with zoom-out features.
\newblock In {\em CVPR}, pages 3376--3385, 2015.

\bibitem{Mundy06}
J.~L. Mundy.
\newblock Object recognition in the geometric era: A retrospective.
\newblock In {\em Toward Category-Level Object Recognition, volume 4170 of
  Lecture Notes in Computer Science}, pages 3--29. Springer, 2006.

\bibitem{Papon15}
J.~Papon and M.~Schoeler.
\newblock Semantic pose using deep networks trained on synthetic {RGB-D}.
\newblock In {\em {ICCV}}, 2015.

\bibitem{Peng15}
X.~Peng, K.~Saenko, B.~Sun, and K.~Ali.
\newblock Learning deep object detectors from {3D} models.
\newblock In {\em {ICCV}}, 2015.

\bibitem{Pepik15}
B.~Pepik, R.~Benenson, T.~Ritschel, and B.~Schiele.
\newblock What is holding back convnets for detection?
\newblock In {\em arXiv:1508.02844}, 2015.

\bibitem{Roberts65}
L.~Roberts.
\newblock Machine perception of {3-D} solids.
\newblock In {\em Ph{D}. Thesis}, 1965.

\bibitem{Russakovsky15}
O.~Russakovsky, J.~Deng, H.~Su, J.~Krause, S.~Satheesh, S.~Ma, Z.~Huang,
  A.~Karpathy, A.~Khosla, M.~Bernstein, A.~C. Berg, and L.~Fei-Fei.
\newblock {ImageNet} large scale visual recognition challenge.
\newblock {\em {IJCV}}, 2015.

\bibitem{Schwing13}
A.~Schwing, S.~Fidler, M.~Pollefeys, and R.~Urtasun.
\newblock Box in the box: Joint {3D} layout and object reasoning from single
  images.
\newblock In {\em {ICCV}}, 2013.

\bibitem{Schwing12}
A.~Schwing and R.~Urtasun.
\newblock Efficient exact inference for {3D} indoor scene understanding.
\newblock In {\em {ECCV}}, 2012.

\bibitem{Silberman12}
N.~Silberman, D.~Hoiem, P.~Kohli, and R.~Fergus.
\newblock Indoor segmentation and support inference from rgbd images.
\newblock In {\em ECCV}, 2012.

\bibitem{SimonyanZ14a}
K.~Simonyan and A.~Zisserman.
\newblock Very deep convolutional networks for large-scale image recognition.
\newblock {\em CoRR}, abs/1409.1556, 2014.

\bibitem{Song14}
S.~Song and J.~Xiao.
\newblock Sliding shapes for 3d object detection in depth images.
\newblock In {\em ECCV}, 2014.

\bibitem{Srivastava14}
N.~Srivastava, G.~Hinton, A.~Krizhevsky, I.~Sutskever, and R.~Salakhutdinov.
\newblock Dropout: A simple way to prevent neural networks from overfitting.
\newblock {\em Journal of Machine Learning Research}, 15, 2014.

\bibitem{Su14}
H.~Su, Q.~Huang, N.~Mitra, Y.~Li, and L.~Guibas.
\newblock Estimating image depth using shape collections.
\newblock {\em ACM Transactions on Graphics (Proceeding of SIGGRAPH)}, 33(4),
  2014.

\bibitem{Su15}
H.~Su, C.~Qi, Y.~Li, and L.~Guibas.
\newblock { Render for CNN: Viewpoint Estimation in Images Using CNNs Trained
  with Rendered 3D Model Views}.
\newblock In {\em {ICCV}}, 2015.

\bibitem{Wang15}
X.~Wang, D.~Fouhey, and A.~Gupta.
\newblock Designing deep networks for surface normal estimation.
\newblock In {\em CVPR}, 2015.

\bibitem{Wu15}
Z.~Wu, S.~Song, A.~Khosla, F.~Yu, L.~Zhang, X.~Tang, and J.~Xiao.
\newblock {3D ShapeNets}: A deep representation for volumetric shape modeling.
\newblock In {\em CVPR}, 2015.

\bibitem{Xiao12}
J.~Xiao, B.~C. Russell, and A.~Torralba.
\newblock Localizing {3D} cuboids in single-view images.
\newblock In {\em {NIPS}}, 2012.

\bibitem{Xie15}
S.~Xie and Z.~Tu.
\newblock Holistically-nested edge detection.
\newblock In {\em ICCV}, 2015.

\end{thebibliography}
}

\end{document}